\documentclass[12pt]{article}

\usepackage[title]{appendix}

\usepackage[english]{babel}
\usepackage{amsmath}
\usepackage{graphicx}

\usepackage{verbatim}
\usepackage[sort]{natbib}
\usepackage{etoolbox}

\usepackage{authblk}

\usepackage[margin=0.9in]{geometry}

\newcommand{\eq}[1]{Eq.\ \ref{#1}}

\newcommand{\nn}{\nonumber}
\newrobustcmd{\disambiguate}[3]{#2~#3}

\title{\Large Bias and population structure in the actuation of sound change}

\author[1]{James Kirby\thanks{j.kirby@ed.ac.uk}}
\author[2,3]{Morgan Sonderegger\thanks{morgan.sonderegger@mcgill.ca}}
\affil[1]{\normalsize School of Philosophy, Psychology and Language Sciences, University of Edinburgh}
\affil[2]{Department of Linguistics, McGill University}
\affil[3]{Centre for Research on Brain, Language, and Music, McGill University}

\date{\vspace{-5ex}}

\begin{document}

\maketitle

\abstract{
Why do human languages change at some times, and not others? We address this longstanding question from a computational perspective, focusing on the case of sound change. 
Sound change arises from the pronunciation variability ubiquitous in every speech community, but most such variability does not lead to change. Hence, an adequate model must allow for stability as well as change. Existing theories of sound change tend to emphasize factors at the level of individual learners promoting one outcome or the other, such as channel bias (which favors change) or inductive bias (which favors stability). Here, we consider how the interaction of these biases can lead to both stability and change in a population setting. We find that population structure itself can act as a source of stability, but that both stability and change are possible only when both types of bias are active, suggesting that it is possible to understand why sound change occurs at some times and not others as the population-level result of the interplay between forces promoting each outcome in individual speakers. In addition, if it is assumed that learners learn from two or more teachers, the transition from stability to change is marked by a phase transition, consistent with the abrupt transitions seen in many empirical cases of sound change. The predictions of multiple-teacher models thus match empirical cases of sound change better than the predictions of single-teacher models, underscoring the importance of modeling language change in a population setting.}

\section{Introduction}





Language changes over time: words come and go, pronunciations shift,
and the structure of sentences mutates, 
such that the `same' language becomes unintelligible to speakers of earlier generations. 
While language change is far from deterministic, it is often strikingly systematic. 
Indeed, it is the regularity of sound-meaning correspondences between words in different languages (e.g. Latin \textit{pedis, pater, pisces} vs. 
English \textit{foot, father, fish}) that licenses hypotheses about a common ancestor. Documenting these sound changes helped to establish linguistics as a scientific discipline in the 18th--19th centuries \citep{jones1788discourse}, resulting in a rich knowledge of \emph{what} types of 
sound changes have occurred in the world's languages \citep{paul1880prinzipien,kuemmel2007konsonantenwandel}.


For almost as long, linguists have asked \emph{why} sound change
occurs---in particular, why particular changes take place, or
\textsc{actuate}, at the time and place they do---a question which has
proven much harder to answer, known as the `actuation problem'
\citep{weinreich1968empirical,baker2008addressing,baker2011variability,garrett2013phonetic}. One
strand of research has emphasized the role of universal phonetic
pressures or \textsc{channel biases} that introduce systematic, potentially asymmetric errors in transmission of a phonetic signal between teacher and learner \citep{ohala1993phonetics,blevins2004evolutionary,moreton2008analytic}. A commonly cited example of a channel bias is \textit{coarticulation}, which causes a speech sound to be produced differently depending on the preceding and following sounds. 
Sound changes such as Germanic $i$-umlaut, whereby low back vowels were fronted and raised when a high front vowel or glide occurred in the following syllable (e.g. Proto-Germanic \textit{gasti} $>$ West Germanic \textit{gesti} `guests', modern German \textit{G\"aste}), have been proposed 
to find their source in this kind of conditioned variation 
\citep{ohala1993phonetics,iverson2003ingenerate,blevins2004evolutionary}. 
%
This leads to a view of actuation as a two-stage process: first, 
an individual learner interprets a coarticulated variant as conventional 
\citep{ohala1981listener}; then, via a process of cultural transmission, the change subsequently spreads throughout the speech community 
\citep{labovprinciples,milroy1980language}.  

While intuitively plausible, important aspects of this model remain to be fully specified. 
First, if 
{channel biases} such as coarticulation are universally active, why are all languages not constantly changing 
\citep{weinreich1968empirical,baker2008addressing,baker2011variability}? It is clear that the presence of bias does not invariably result in change: for instance, even while umlaut was spreading throughout the West Germanic languages, it did not affect Gothic \citep{cercignani1980alleged}. 
An adequate model of sound change must 
therefore 
also account for the possibility, even ubiquity, of stable variation at the level of the speech community.
One explanation for stability would be the existence of (possibly domain-general) \textsc{inductive biases} guiding human inferences, which may facilitate or inhibit the learning of certain types of structures or patterns \citep{briscoe2000grammatical,reali2009evolution,kalish2007iterated,griffiths2007language}. Inductive biases have been proposed that favour phonetically-motivated hypotheses about phonological patterns over phonetically arbitrary ones (e.g.\ \textit{substantive biases}: \citealp{moreton2008analytic,steriade2008phonology,wilson2006learning})\footnote{The terms `inductive bias', `analytic bias' and `learning bias' are often used interchangeably in this literature; see e.g. \cite{moreton2008analytic,moreton2012structure}.} or which promote the stability of 
existing phonetic category structures over the creation of new ones (e.g.\ \textit{categoricity biases}: \citealp{pierrehumbert2001exemplar,wedel2006exemplar}). However, if such preferences are strong enough to counteract channel bias, then how can change ever occur? Finally, when change does diffuse throughout a speech community, it often occurs suddenly following a period of prolonged stability \citep{labovprinciples,kroch1989reflexes}. What types of constraints on transmission and learning might interact to produce this type of rapid shift from one stable state to another?

In this paper, we address
these questions by modeling the acquisition and propagation of a phonetic parameter in a population setting. Our goal is a model that predicts both stability and change in the presence of biases promoting the other outcome, and in which small changes in the magnitude of bias produces a sudden and nonlinear change from one stable state to another. 	
Because such questions about language change are difficult to address empirically, we approach this problem from the perspective of computational and mathematical modeling
, drawing on a large body of previous work in this tradition \citep{kirby2013role,blythe2012scurves,boyd1985culture,cavalli1981genetic,niyogi1997evolutionary,komarova2001evolutionary,kroch1989reflexes,niyogi2006computational,sonderegger2010combining,griffiths2013effects,pierrehumbert2001exemplar,wedel2006exemplar,griffiths2007language,kirby2007innateness,reali2009evolution,smith2003iterated,dediu2009genetic,burkett2010iterated,niyogi2009proper,smith2009iterated}.  
Our approach differs crucially from previous work in two respects. First, while models of language change often 
frame the learner's task as choosing between competing discrete
variants
\citep{yang2000internal,kroch1989reflexes,baker2011variability,niyogi2006computational,wang2004computational,sonderegger2010combining}, 
a key part of learning the sound pattern of a language is learning
distributions over continuous phonetic parameters, such as vowel
formants \citep{vallabha2007unsupervised}.
Second, in most existing models 
that have considered continuous parameters, 
 change \textit{only} and \textit{always} occurs in the presence of a channel bias \citep{pierrehumbert2001exemplar,wedel2006exemplar,baker2008addressing,kirby2013role}. 
Here, we propose a model in which both stability and change of a continuous parameter are possible in the presence of channel bias.

By stability, we are referring to the structure of the stationary distribution of the continuous parameter in the population. Might stability at the population level have its roots in the inductive biases of individual learners?  This seems plausible given work on the dynamics of cultural transmission showing that 
the distribution of a cultural trait evolves linearly to a unique stationary state that reflects the structure of learners' prior 
\citep{griffiths2007language,kirby2007innateness,reali2009evolution}. However, while this result holds for chains of single teacher-learners, in general 
the dynamics become nonlinear as the population structure becomes more complex \citep{dediu2009genetic,burkett2010iterated,niyogi2009proper,smith2009iterated}, opening up the possibility of very different outcomes---such as stability and change---from similar initial conditions \citep{niyogi2009proper}. 
In what follows, we  
thus consider population structures of increasing complexity, and assess our models on the basis of their ability to explain how a nonlinear transition from stable variation to sound change could occur.

\section{Model}
\label{modelSec}


We consider a scenario in which each agent may 
(a) function as a \textit{learner}, receiving input from other agents and applying a learning algorithm to this input in order to learn a probability distribution over how a continuous parameter is realised, and (b)
function as a \textit{teacher}, generating data from this distribution for other learners. Within this framework, there are many assumptions one could make about each of these actions. 
Here we 
consider variants on a simple supervised learning scenario, where all that needs to be learned is a distribution over a single phonetic dimension, parametrized by a single continuous parameter. For concreteness, our exposition follows the example of umlaut described in the Introduction, but the basic results are applicable to the learning of a single continuous parameter more generally.

\subsection{Linguistic setting}\, 
We assume that 
speech sounds have been organised into discrete segments, and that
the learner has access to the complete segmental inventory.
We consider here a simple language with the lexicon $\Sigma = \{ \text{V}_1, \text{V}_2, \text{V}_{12} \}$, where \text{V}$_{12}$ represents V$_1$ in the context of V$_2$. 
$\text{V}_1$ and $\text{V}_2$ can be thought of as the vowels /a/ and /i/ in isolation, and $\text{V}_{12}$ as  /a/ in a context where it is coarticulated (raised) towards /i/. 

Tokens are represented by their first formant (F1) value, an acoustic
measure of vowel height \citep{hillenbrand1995acoustic}.\footnote{Note
  that F1 is inversely related to physical tongue height, so F1 is
  lower for /i/ than for /a/.}  . We assume that the F1 distributions
for
V$_1$ and V$_2$ are 
normal  ($N(\mu_a, \sigma_a^2)$, $N(\mu_i, \sigma_i^2)$), are known to all learners, are the same for all learners, and do not change over time.\footnote{This could be taken to mean that learners in generation $t$ receive a very large number of V$_1$ and V$_2$ examples, and 
  learn these distributions perfectly from generation $t-1$.} The distribution of V$_{12}$ is normal, with (fixed) variance as for V$_1$ and a mean we denote by $c$:
\begin{equation} 
\text{V}_{12} \sim N(c, \sigma_a^2)
\label{threeDists}
\end{equation}
We will sometimes refer to $\text{V}_{12}$ (or equivalently, $c$, which determines the distribution of $\text{V}_{12}$) as the \textit{contextual variant}.


In addition, we assume that productions of V$_{12}$ are subject to a coarticulatory channel bias corresponding to the general tendency in speech production to over- or undershoot articulatory targets based on speech context \citep{pierrehumbert2001exemplar,lindblom1983economy}. 
We take this bias to be normally distributed with mean $-\lambda$ (because V$_2$ has lower F1 than V$_1$)
 and variance $\omega^2$, and to be applied i.i.d.\ to each vowel token.  Thus, the actual productions of V$_{12}$ by a teacher with contextual variant $c$ follow the distribution
\begin{equation}
F1 \sim N(c - \lambda, \sigma_a^2 + \omega^2)
\label{v12ProdBias}
\end{equation}

\subsection{Learning and evolution}

%

We assume agents are divided into discrete generations of size $M$.  Each learner in generation $t+1$ receives $n$ examples of V$_{12}$ (distributed according to \eq{v12ProdBias}) from one or more teachers from generation $t$.
The learner's task is to infer $c$ by application of some learning algorithm.

We assume that 
learners apply a learning algorithm which is `rational', in the sense that they assume that their learning data was generated i.i.d.\ according to \eq{threeDists},
and estimate the most probable value of $c$.  Results are presented below for three  learning algorithms (\emph{Naive learning models}, \emph{Simple prior models}, \emph{Complex prior models}) corresponding to different assumptions about learners' inductive biases. Here, we specifically model the effect of a categoricity bias, operationalised as a prior over values of $c$.

For each case, we consider three population structures (Fig.\ \ref{pops}), corresponding to the number of teachers $m$ in generation $t$ each learner in generation $t + 1$ receives her learning data from: $m=1$, $m=2$, and $m=\text{all}$ (equivalently, $m=M$).  These three values are chosen as representative for understanding the dynamics when \emph{any} number $m$ of teachers is assumed, which we are interested in in light of previous computational studies highlighting the differences between single- and multiple-teacher scenarios in language evolution \citep{dediu2009genetic,burkett2010iterated,niyogi2009proper,smith2009iterated}.  The single-teacher case corresponds most closely to the population structure considered in `iterated learning' models of language evolution \citep[e.g.][]{smith2003iterated,griffiths2007language,kirby2007innateness,reali2009evolution}.\footnote{Although our single-teacher scenario is closest to that considered in iterated learning models, there remain important differences, as discussed in Appendix \ref{app:modelSec}.}  The all-teachers case corresponds to the population structure usually assumed in dynamical systems models of language change
\citep[e.g.][]{niyogi1997evolutionary,niyogi2006computational,sonderegger2010combining}.  The two teacher-case is representative of all $m$ between 2 and $M - 1$, because the dynamics turn out to be extremely similar for any $m>1$, as we show below.
We assume throughout that the $m$ teachers are chosen uniformly from teachers in the previous generation, with replacement.

Considering the ensemble of $M$ teachers in generation $t$, the state of the population at $t$ can be characterized by the random variable $C^t$, whose distribution describes how likely different values of $c$ are.  Similarly, the values of $c$ learned by the $M$ learners in generation $t+1$ can be characterized by $C^{t+1}$.
For simplicity, we assume that $M$ is infinite.  The evolution of the distribution of $c$ is then deterministic, 
making its behavior more easily analyzed as a dynamical system.
This and several other
 aspects of our modeling framework (e.g.\ discrete generations)
are shared with previous dynamical systems models of language change considering \emph{discrete} variants \citep{niyogi1997evolutionary,niyogi2006computational}.



\begin{figure*}
\centering
\includegraphics[width=1.0\linewidth]{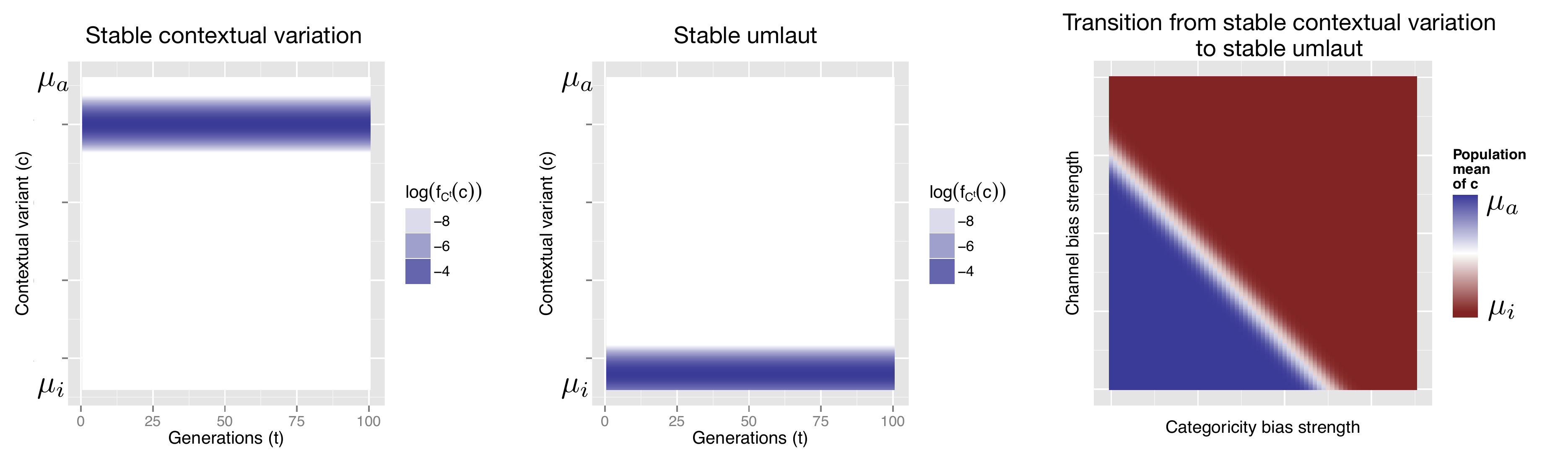}
\caption{\label{schematic} 
Schematic of possible distributions of the contextual variant ($c$) in the population over time, and posible dependence of the mean value of $c$ on system parameters. (A) Stable contextual variation: the distribution of $c$ in the population is stable over time, and its mean is closer to $\mu_a$ than to $\mu_i$. (B) Stable umlaut: the distribution of $c$ is stable over time, and its mean is near $\mu_i$.  (C) Nonlinear transition from stable variation to stable umlaut. The mean of the stable population-level distribution of $c$ depends on two parameters: the strengths of the coarticulatory channel bias and the categoricity bias.  For most parameter values, there is stable contextual variation or stable umlaut; a nonlinear transition from one to the other occurs when a boundary in parameter space is crossed.  
}
\label{fig:schematic}
\end{figure*}

\begin{figure*}
\centering
\includegraphics{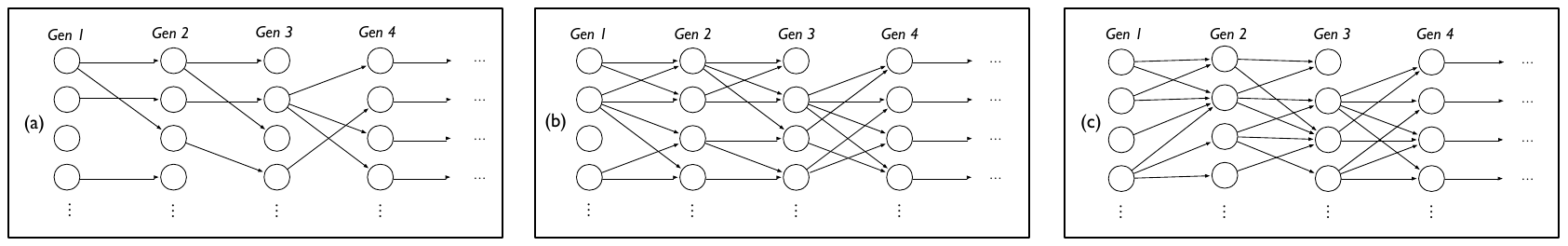}
\caption{\label{pops} 
Three types of population structure are considered in our models: (a)  Single-teacher scenario. Each learner in generation $t+1$ receives all her learning data from a single randomly-chosen teacher in generation $t$. (b) Multiple-teacher scenario (two teachers).  Each data point comes from one of two teachers with equal probability. (c) Multiple-teacher scenario ($M$ teachers).  Each data point comes from a random teacher with equal probability.  In (b)--(c), teachers are chosen uniformly at random from teachers in generation $t$ (with replacement).  In all cases, lines of descent may be pruned, i.e. some teachers may not provide data to any learners in the following generation.}
\label{fig:learning}
\end{figure*}

Given a choice of learning algorithm, population structure, and channel bias, we seek to characterize the evolution of the distribution of $c$, and determine to what extent it satisfies our modeling goals: stability in the presence of channel bias, change in the presence of categoricity bias, and a nonlinear shift from one stable state (where the distribution of $c$ does not change over time) to another.
We are especially interested in two types of stable state ---\emph{stable contextual variation}, where the mean value of $c$ in the population is nearer to $\mu_a$ than to $\mu_i$, and \emph{stable umlaut}, where this mean is near $\mu_i$. Fig.\ \ref{fig:schematic} exemplifies what the distribution of $c$ in the population over time could look like in both cases, as well as one possible way in which a nonlinear shift from stable contextual variation to stable umlaut could occur as system parameters are varied.\footnote{Note that in the right panel of Fig.\ \ref{fig:schematic}, what is important for our purposes is the nonlinearity of the transition between stable contextual variation and stable umlaut, rather than the shape of the boundary in parameter space across which the transition occurs (which could be any curve, rather than the line shown in Fig.\ \ref{fig:schematic}).}



In the remainder of the paper, we first consider the simplest case, where
learners have no prior on values of $c$ (\emph{Naive learning models}));
we then consider the effects of introducing
different types of categoricity bias into the learning algorithm
(\emph{Simple prior models} and \emph{Complex prior models}),
and conclude by
discussing our results.

For each class of model (naive, simple prior, complex prior), we are
interested in the evolution of the distribution of $c$,
for which there is no general analytic solution. For the naive
learning models and simple prior models, we consider
how the mean and variance of this distribution change over time, which can be derived analytically using techniques familiar from the 
 cultural evolution literature 
 \citep{boyd1985culture,griffiths2013effects} and dynamical systems
 models of language change in discrete variants; derivations for all
 analytic results are
are given in
Appendix \ref{app:naiveSec}--\ref{app:simplePriorSec}.  For the complex prior
 models,
we proceed by simulation.

\section{Naive learning models}
\label{naiveSec}

We first consider maximum-likelihood (ML) learners, who are `naive'
in the sense of having no prior over 
$c$, and simply choose the  value of $c$ under which the likelihood of the data (according to \eq{threeDists}) is highest.
%


In the case where each learner in generation $t+1$ receives all $n$ examples from a single teacher, 
%
%
the evolution of the mean and variance of $c$ are:
 \begin{align}
   E[C^{t+1}] & = E[C^t] - \lambda  \label{mod11Mean} \\
   \text{Var}(C^{t+1}) &= \frac{\sigma_a^2 + \omega^2}{n}+ \text{Var}(C^t) \label{mod11Var}
\end{align}
Thus, when there is coarticulation ($\lambda > 0$), the mean of the contextual variant decreases in every generation by an amount equal to the mean amount of channel bias; if there is no coarticulation ($\lambda = 0$), the mean stays the same over time.  Regardless of the value of $\lambda$, however, the inter-speaker variability in the realization of the contextual variant
 increases without bound over time.
Next, consider the case where each learner receives all examples from two teachers. The evolution of the mean 
of the contextual variant ($c$) in this case is again described by \eq{mod11Mean}, 
while the variance now rapidly converges to a fixed point, regardless of the initial distribution: 
\begin{equation}
  \text{Var}(C^{t}) \to \frac{2 (\sigma_a^2 + \omega^2)}{n-1}
  \label{mod13Var}
\end{equation}
Thus, the mean value of $c$ 
 decreases without bound over time ($\lambda > 0$) or stays constant
 ($\lambda=0$), while the variance quickly stabilizes, in contrast to
 the single-teacher case.

 In fact, it can be shown that the dynamics are similar for any case where $m > 1$: 
the mean of $c$ is described by 
\eq{mod11Mean}, while
 its variance moves towards a fixed point. The larger $m$ is, the
 smaller this stable variance is (Appendix \ref{app:naiveMultipleSec}). In the
 limiting case where $m=M$ (Fig.\ \ref{pops}c), the variance converges
 to: 
\begin{equation}
  \text{Var}(C^{t}) \to \frac{\sigma_a^2 + \omega^2}{n-1}
\label{mod12Var}
\end{equation}

\begin{figure}
\centering
\includegraphics[width=0.75\textwidth]{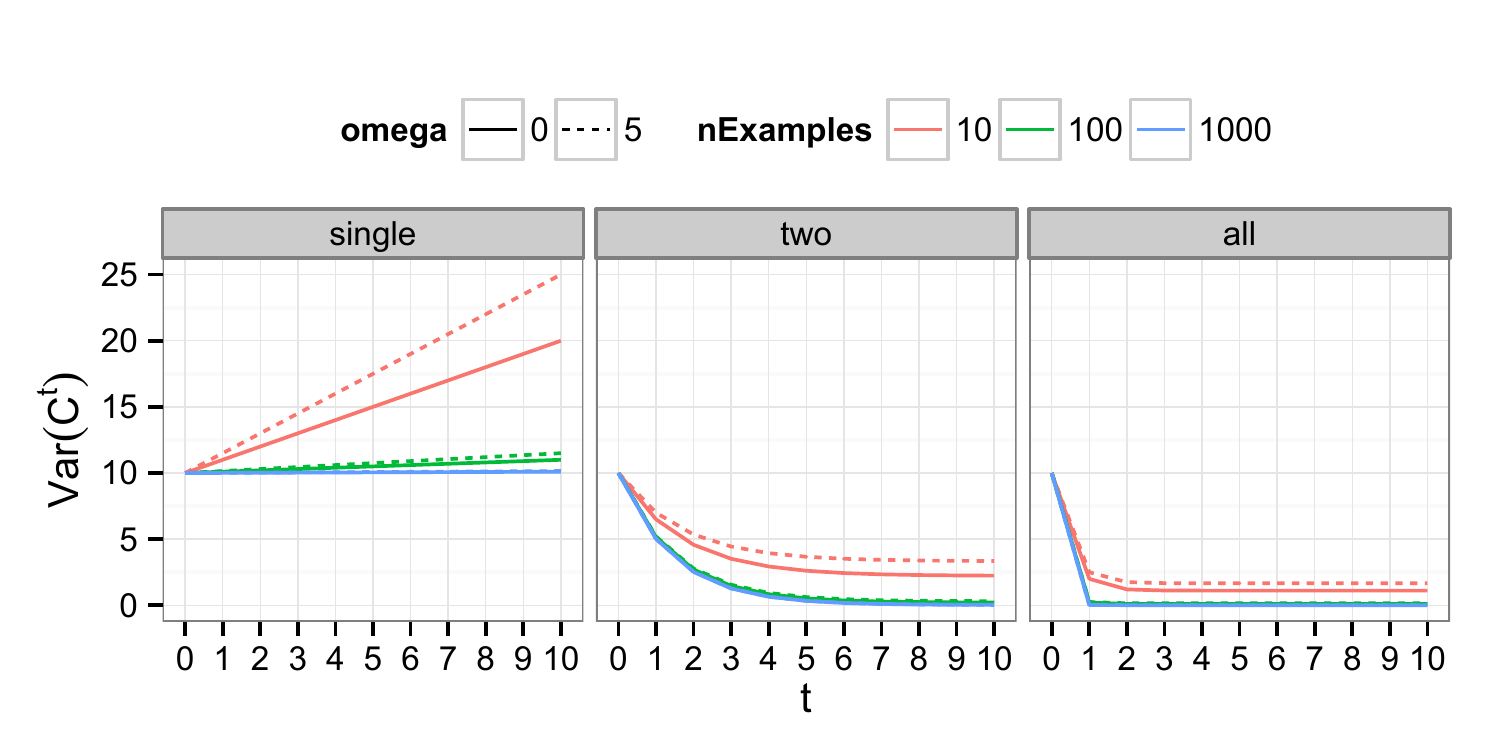}
\caption{\label{vars}Evolution of population variance Var($C^t$) for different numbers of training examples, assuming $\sigma_a^2=60$, $\omega = 0$ or $\omega=5$, and $m$ = 1, 2, or $M$. In the single-teacher ($m=1$) setting, the variance decreases without bound over time, while for two or more teachers, it rapidly stabilizes.}
\end{figure}

The evolution of the variance in the single-teacher, two-teacher, and
all-teacher cases are illustrated in Fig.\ \ref{vars}.

\paragraph{Summary}
In the naive learning models, if speakers do not coarticulate, the mean realization of V$_{12}$
in the population remains constant over time, regardless of the number of teachers. This is empirically inadequate, as it predicts change from
stable contextual variation to stable umlaut
to be impossible.  In the presence of any channel bias ($\lambda > 0$),
the mean 
of $c$ in the population steadily increases over time, again regardless of the number of teachers.  In this case, change from stable contextual variation to stable umlaut is not possible, because \emph{stability} is not possible, in the sense of a distribution of $c$ which does not change over time.  This problem is even worse for the single-teacher model, where the variance of $C^t$ in the population is predicted to steadily increase.  As far as we are aware, a permanently unstable and unstructured distribution of population-level variation in phonetic realization is uncharacteristic of speech communities.

\section{Simple prior models}
\label{simplePriorSec}

\begin{figure}
  \centering
  \includegraphics[width=0.8\linewidth]{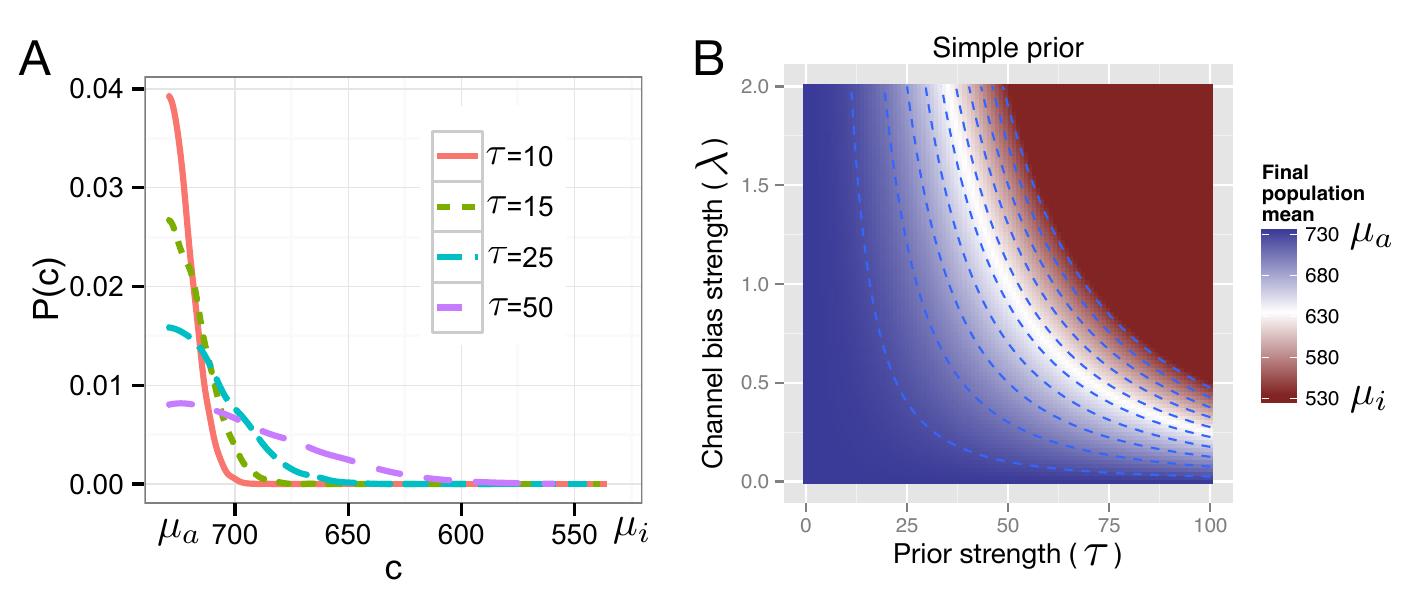}
  \caption{\label{simplePriorFig} 
    Simple prior models setup and results, with $\mu_i = 530$, $\mu_a = 730$, $n=100$, $\sigma_a = 50$. \emph{(A)} Prior distribution over $c$ 
    ($N(\mu_a, \tau^2)$) for values in $[\mu_i, \mu_a]$.  The parameter $\tau$ controls the prior strength, with 
    values closer to 0 corresponding to a greater preference for values of $c$ near $\mu_a$. 
    \emph{(B)} 
    Final population mean of $c$ as a function of channel bias ($\lambda$) and prior strength ($\tau$), assuming the minimum value is $c = \mu_i$ (for comparability with Fig.\ \ref{complexPriorFig}B).  The final mean does not depend on the number of teachers or the starting state of the population, and changes gradually as $\lambda$ and $a$ are changed.}
\end{figure}

The main problem with the naive learning models change becomes inevitable once channel bias is introduced. This problem has 
led to criticism of theories of sound change that rely on the accumulation of incremental change \citep{weinreich1968empirical,baker2008addressing,baker2011variability}
. However, the inevitability of change in these models is not simply a function of the channel bias itself, but also because there is no force acting to counteract the bias. Perhaps the simplest type of countervailing force would be to assume that learners have a prior categoricity bias over $c$ against values away from $\mu_a$.
In particular, consider a simple gaussian prior centered at $\mu_a$ with variance $\tau^2$ (Fig.~\ref{simplePriorFig}A), a type that has previously been considered in work on the evolution of a continuous parameter (\citealp{griffiths2013effects}: we compare this study's results to ours in \ref{singleTeacherSec}).

Learners receive $n$ examples in the same way as in the naive learning models, but now their 
knowledge about  contextual variation is probabilistic: a given learner begins with a prior distribution
on how likely different amounts of contextual variation are \emph{a
  priori}, which is updated to a posterior distribution based on her
data, assuming that the distribution of the data given $c$ is given by
\eq{threeDists}.  She then takes the maximum a-posteriori (MAP)
estimate as her point estimate $\hat{c}$ of the contextual
variant.\footnote{The other common strategy for obtaining a point
  estimate of a posterior distribution, taking the expected value,
  turns out to be equivalent (Appendix \ref{app:simplePriorSec}).}
As for the naive learning models, we consider the evolution of the mean and variance of $C^t$.

Regardless of the number of teachers, 
the mean of $C^t$ rapidly moves towards a fixed point, namely:
\begin{equation}
E[C^t] \to  \mu_a - \lambda n \frac{\tau^2}{\sigma_a^2} \quad \text{(1, 2, $\ldots$ $M$ teachers)}
\label{simpleMean}
\end{equation}
Thus, the stronger the prior bias against contextual variation
there is (smaller $\tau$), the smaller the eventual mean degree of contextual variation
in the population, but increasing the strength of the channel bias (larger $\lambda$) has the opposite effect (Fig.\ \ref{simplePriorFig}B).

As in the naive learning models where $m \geq 2$, the variance of $C^t$  always rapidly moves towards a fixed point for all types of population structure. 
The formulas for these fixed points depend on $\sigma_a$, $\omega$, $\tau$, and $n$.  To get a sense of their essential properties, we write them in a form which assumes $n \gg 0$:
\begin{align}
\text{Var}[C^{t+1}] & \to \tau^2 \frac{K}{2} - \sigma_a^2 \frac{K}{4n} + O(\frac{1}{n^2})  & \text{(one)} \label{simpleVar1} \\
    & \to  \sigma_a^2 \frac{2K}{n} + O(\frac{1}{n^2})  & \text{(two)}  \label{simpleVar2} \\
  & \to \sigma_a^2 \frac{K}{n} + O(\frac{1}{n^2})  & \text{($M$)} \label{simpleVar3} 
\end{align}
where $K = (1 + \omega^2/\sigma_a^2)$ and $O(\frac{1}{n^2})$ denotes a constant divided by $n^2$. 
While the variance 
always stabilizes over time, even for the single-teacher case, comparing Eqs.\ \ref{simpleVar1}--\ref{simpleVar3}
shows that (for large enough $n$) just as for the naive learning models, the larger the number of teachers, the smaller the eventual amount of population-level variability in $c$.  

\paragraph{Summary}
The qualitative evolution of $c$ in simple prior models is the same regardless of the magnitude of channel bias (including when $\lambda = 0$): both the mean and variance of the realization of V$_{12}$
 in the population 
always move to a stable value. In the limit of large $n$, the stable variance shows an important qualitative difference that depends on the number of teachers: 
while convergence to a form reflecting prior is seen in the single-teacher scenario, the stable value in scenarios with two or more teachers does not directly reflect the prior ($\tau$ is not a term in Eqns. \ref{simpleVar2}--\ref{simpleVar3}, cf. \citealp{griffiths2013effects}).

The simple prior models allow for stable contextual variation at a value that depends on the relative strengths of the channel and categoricity biases. 
However, these models are in some sense \emph{too} stable: 
because stability depends on particular values of the system parameters, in order for a change to `go to completion' (i.e., to stable umlaut)
the system parameters would need to be continually changing in each generation---implying that each generation coarticulates more than the previous, has a weaker categoricity bias, or both. While this is ultimately an empirical question, it seems to us useful to start from the assumption that the effects of purportedly universal biases do \textit{not} change steadily over time. In this sense, the simple prior models are inadequate in that there is no threshold in the system parameters triggering rapid movement to stable umlaut.
 


\section{Complex prior models}
\label{complexPriorSec}

The simple prior is indeed a type of categoricity bias, but one that is asymmetrically biased entirely toward one of the two pre-existing categories. 
Here, we consider the ramifications of relaxing this assumption, assuming instead that learners have a complex prior
which weights values of $c$ near \textit{both} $\mu_a$ or $\mu_i$ higher than values in between:
\begin{equation}
  P(c) \propto \left[ a(\mu_a - \mu_i)^2 + (c - (\mu_a + \mu_i)/2)^2 \right]
  \label{post3}
\end{equation}
The strength of this prior is controlled by $a$: as $a \to 0$, values of $c$ near $\mu_a$ and $\mu_i$ are maximally preferred
(Fig.\ \ref{complexPriorFig}A).

We assume the learner takes the MAP estimate $\hat{c}$ over the range $[\mu_i, \mu_a]$.  
Unlike in previous models, the mean and variance of $C^t$ cannot be determined analytically, and we thus proceeded by simulation to 
%
%
determine the evolution of the distribution of $C^t$ over time in this
case.  Technical details of these simulations are given in \ref{simulationsSetupSec};
here we describe the basic setup of the simulations, and their results.

The simulations described below consider the evolution of a
population that starts with a mean realization of V$_{12}$ similar to
V$_{1}$ ($C^1 \sim N(\mu_a - 10, \sigma_a^2)$), in order to determine
whether both stable contextual variability and change to stable umlaut
are possible in this model.
Of interest is how the strength of the prior ($a$) and the
coarticulatory channel bias ($\lambda$) affect the evolution of the distribution
from this starting point, which we examine for the same three
population structures as in previous models.  We first examine the
evolution of the distribution of $C^t$ over time (which we refer to as
the \emph{trajectory} of $C^t$) for particular values of $a$,
$\lambda$, and $m$ (\emph{Trajectories of $C^t$}),
 then examine how the final mean of the
distribution of $C^t$ depends on these three parameters (\emph{Final mean of $C^t$}).




\subsection{Trajectories of $C^t$: examples}
\label{trajectoriesCtSec}

\begin{figure}
  \includegraphics[width=\linewidth]{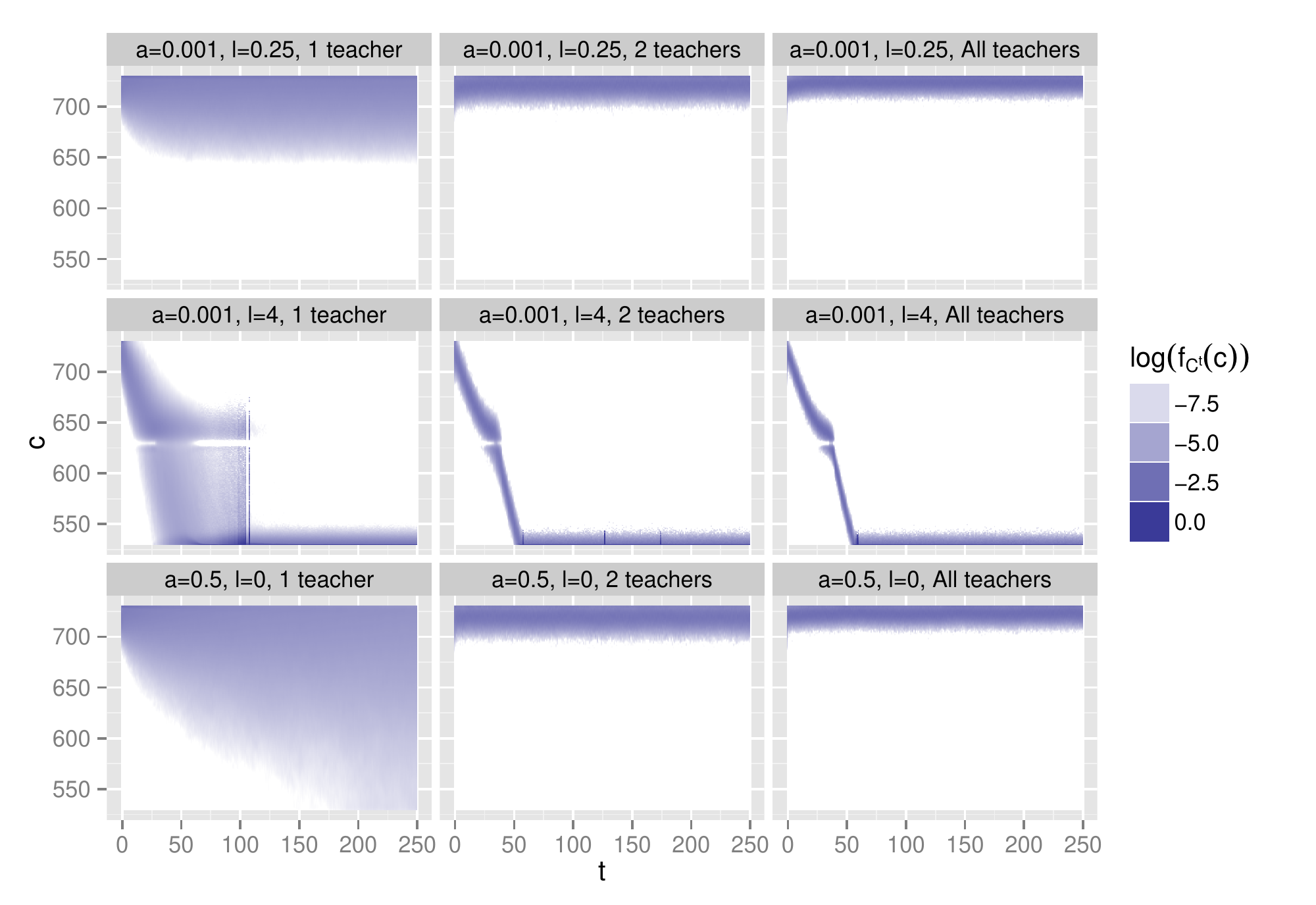}
  \caption{Evolution of PDF of $C^t$ (thresholded
    at $f_{C^t}(c) = 0.0001$) from a starting distribution of $C^1
    \sim N(\mu_a - 10, 10)$, divided into columns by the number of
    teachers ($m$), for values of $a$ and $\lambda$ which result in
    stable contextual variation (top row), change to stable umlaut
    (middle row), and similar behavior to the naive prior models
    (bottom row).   $\mu_a = 730$, $\mu_i = 530$, and other parameters
    listed in  Appendix \ref{simulationsSetupSec}.}
\label{complexPriorTrajectoriesFig1}
\end{figure}

\begin{figure}
\includegraphics[width=\linewidth]{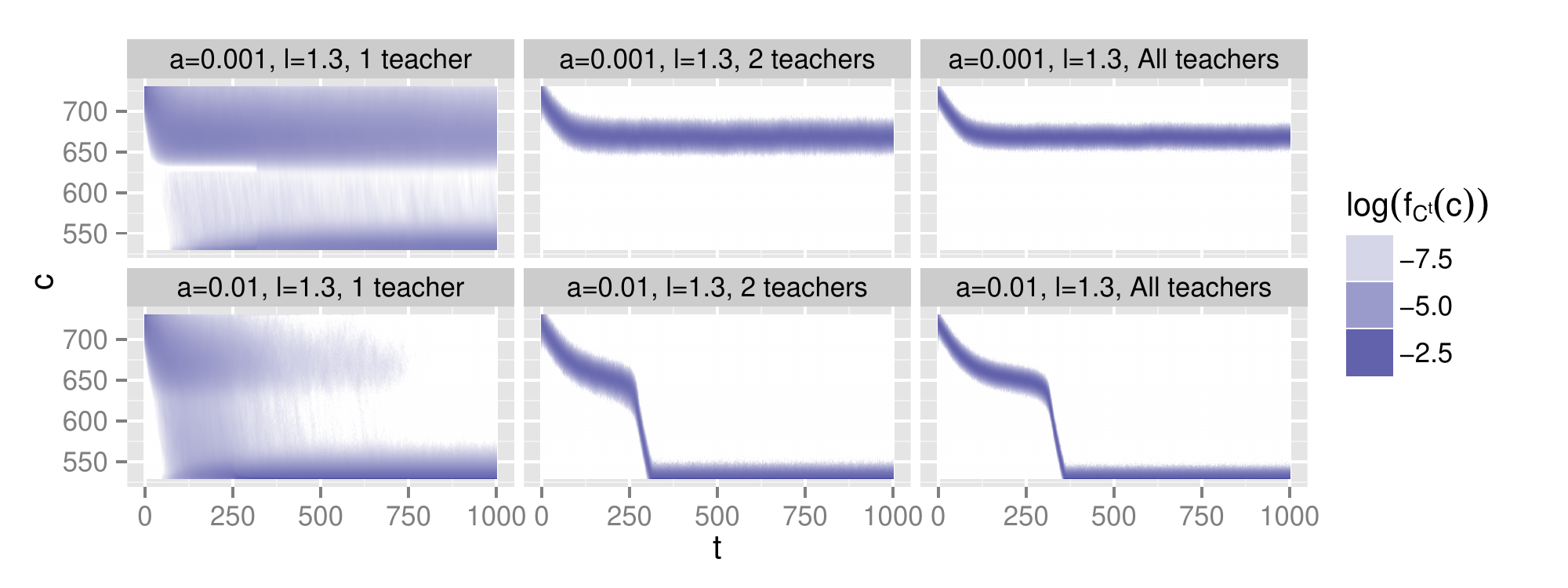}
  \caption{Evolution of PDF of $C^t$ (thresholded at $f_{C^t}(c) = 0.0001$) from a starting distribution of $C^1 \sim N(\mu_a - 10, 10)$, divided into columns by the number of teachers ($m$), for values of $a$ and $\lambda$ which give qualitatively different behavior for $m=1$ and $m>1$. $\mu_a = 730$, $\mu_i = 530$, and other parameters
    listed in  Appendix \ref{simulationsSetupSec}.} 
\label{complexPriorTrajectoriesFig2}
\end{figure}


We show some qualitatively different ways in which the
distribution of $C^t$ can evolve, by examining the trajectories of
$C^t$ beginning from $C^1 \sim N(\mu_a - 10, 10)$, for particular
values of $a$, $\lambda$, and $m$, stopping each simulation when
$t=1000$.  (It is visually clear from the results of these
simulations, shown in Figs.\
\ref{complexPriorTrajectoriesFig1}--\ref{complexPriorTrajectoriesFig2},
that the distribution of $C^t$ is no longer changing by this point,
i.e.\ has reached a stable state.)

To get a sense of the effect of the joint effect of the complex prior
and channel bias on the dynamics of $c$,
we first
 consider trajectories for three limiting cases, shown in Fig.\ \ref{complexPriorTrajectoriesFig1}:

\begin{itemize}
\item \emph{Case 1: strong prior, weak channel bias} (top row:
  $a=0.001$, $\lambda = 0.25$): For a sufficiently strong prior
  relative to the strength of the channel bias, contextual
  variation is stable over time (for 1, 2, all teachers).The stable
  variance of the distribution is much larger for $m=1$ than for
  $m>1$, and is slightly larger for $m=2$ than for $m=M$.

\item \emph{Case 2: strong prior, strong channel bias} (middle
  row: $a=0.001$, $\lambda = 4$): 
  For a sufficiently strong channel bias relative to the prior strength, change to stable umlaut rapidly occurs (for 1, 2, all teachers). The
  transition is slightly faster for $m>1$ than for $m=1$.  
  
\item \emph{Case 3: weak prior, weak channel bias} (bottom row:
  $a=0.5$, $l=0$): In the single-teacher case, the variance rapidly
  spreads, and all values of $c$ become roughly equiprobable. For more
  than one teacher, the mean changes little and the variance rapidly
  stabilizes, with the value of the stable variance is slightly larger for
  $m=2$ than for $m=M$.  These behaviors are similar to the analogous
  naive learning models, as expected given that a sufficiently weak
  prior is effectively flat.
\end{itemize}

In Cases 1--3, the evolution of $C^t$ looks qualitatively similar for $m=1$, $m=2$, and $m=M$, with a significantly larger variance of $C^t$ at each time point for the $m=1$ case.  
However, there is also a range of $(a, \lambda)$ parameter space where the evolution of $C^t$ looks qualitatively different depending on the number of teachers.  Fig.\ \ref{complexPriorTrajectoriesFig2} shows two ways in which this can happen:
\begin{itemize}
\item  \emph{Case 4: strong prior, medium channel bias} (top row:
  $a = 0.001, \lambda = 1.3$): Regardless of the number
  of teachers, the stable state of the population shows stable
  contextual variation, in the strict sense defined above, that the mean of $c$ in the population is closer to $\mu_a$ than to $\mu_i$, but this is realized in qualitatively different ways for $m=1$ and $m>1$.   In the single-teacher case, the distribution of $C^t$ reflects the (strong) prior, in the sense that some individuals have values of $c$ near $\mu_a$ (contextual variation) and some have values of $c$ near $\mu_i$ (umlaut), with a gap in between.  That is, change to umlaut has `gone through' for some individuals, but not others. In contrast, in the multiple teacher cases, the distribution of $C^t$ becomes tightly clustered around  the population mean (which is nearer to $\mu_a$ than to $\mu_i$). 

\item \emph{Case 5: medium prior, medium channel bias} (bottom row: $a = 0.01, \lambda = 1.3$):  In this case, the channel bias is kept at the same value, but the prior is weakened sufficiently that change to stable umlaut eventually occurs, regardless of the number of teachers. However, the trajectory of $C^t$ looks qualitatively different depending on the number of teachers. For $m=1$, the population contains two types of individuals---those with values of $c$ near $\mu_a$, and those with values of $c$ near $\mu_i$---and the proportion of the second type becomes greater over time, until the whole population has $c$ near $\mu_i$.  For $m>1$, individuals have values of $c$ tightly clustered around the population mean, which steadily changes from near $\mu_a$ to near $\mu_i$ over time.
\end{itemize}

In Cases 4--5, it is again the case (as in Cases 1--3)
that the two-teacher and
$M$-teacher cases look very similar, with a slightly larger variance
of $C^t$ when $m=2$.

\begin{figure*}
  \includegraphics[width=\linewidth]{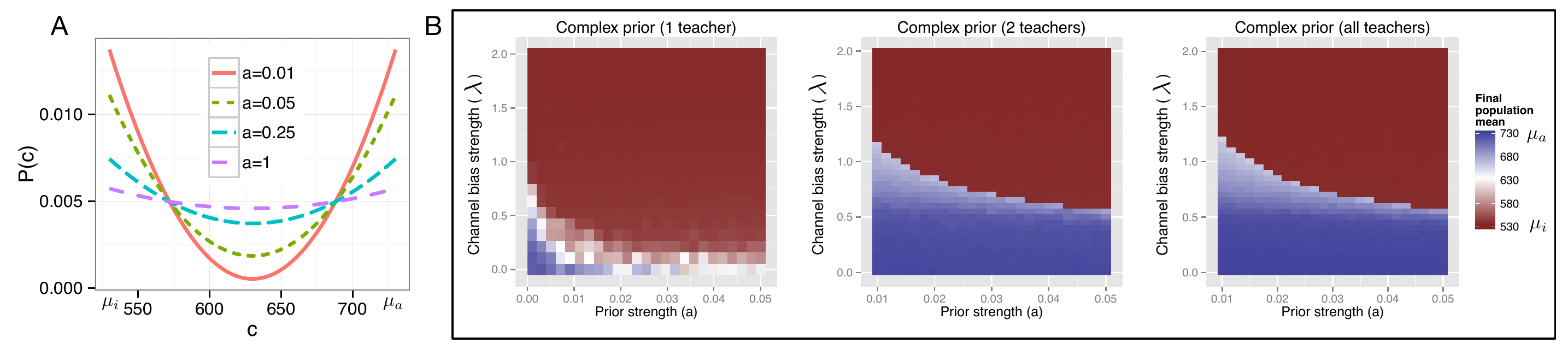}
\caption{\label{complexPriorFig}
    Complex prior models setup and results, with $\mu_a = 730$ and
    $\mu_i = 530$, and other parameters listed in Appendix \ref{simulationsSetupSec}. \emph{(A)} Prior distribution over $c$ (\eq{post3}) for 
    values in $[\mu_i, \mu_a]$. The parameter $a$ controls the strength of the prior, with values nearer to 0 corresponding to a greater preference for values of $c$ near either endpoint. 
    \emph{(B)} Final population mean of $c$, beginning from the same starting state, 
as a function of channel bias ($\lambda$) and prior strength ($a$). The final mean of $c$ depends on the number of teachers (1 vs. 2+), and changes nonlinearly as $\lambda$ and $a$ are changed. In particular, for 2+ teachers there is a bifurcation: once $\lambda$ is large enough relative to $a$, rapid change to stable umlaut occurs.}
  
\end{figure*}


Of the trajectories considered above, Cases 1--2 are particularly
important: they show that
both stable contextual variation and stable umlaut are possible, as $a$ and $\lambda$ are varied.  In particular, it is possible to get change to stable umlaut in the presence of a strong categoricity bias---which was not possible in the simple prior model---as well as stable contextual variation near $\mu_a$ in the presence of channel bias. These outcomes are two of our modeling goals.  We now consider how the final state of the population depends on prior strength and channel bias, as $a$ and $\lambda$ are varied between these limiting cases, to get a sense of whether the complex prior model meets our final modeling goal: a threshold in the system parameters ($a$ and $\lambda$) which triggers rapid movement to stable umlaut.



\subsection{Final mean of $C^t$ as a function of system parameters}
\label{finalMeanSec}

Fig.~\ref{complexPriorFig}B shows the final mean of $c$ in the population as $\lambda$ and $a$ are varied. In the single-teacher case (panel 1),
stable contextual variation 
is possible only for the strongest priors or when $\lambda=0$. As the
strength of the prior is relaxed, the population mean comes to rest
either in an intermediate state, or near $\mu_i$ (i.e.\ stable
umlaut).   The distribution of $C^t$ in an intermediate state often
corresponds to Case 4 above: 
individual learner's means are not tightly clustered around the population mean, but reflect the prior in the sense that some individuals are stable near one endpoint ($c=\mu_a$) and some near the other ($c = \mu_i$), corresponding to an empirical population in which a change has gone through for some speakers but not for others.

In multiple-teacher scenarios (panels 3-4), the results are quite different. There is a range of values of prior strength and channel bias which give stable contextual variation.
However, for a given $a$, as $\lambda$ is increased past a critical value, there is a rapid shift of the population to a stable state where most learners have umlaut ($c \approx \mu_i$).  That is, there is a \emph{bifurcation} where the strength of coarticulation has overcome the stabilizing affect of the prior. 
When this happens, the population mean rapidly moves towards the other category mean and stabilises.
Panels 3--4 also illustrate the tradeoff between categoricity and channel biases: for a stronger prior, the critical value of $\lambda$ increases (i.e., the degree of coarticulation needed to overcome the prior is greater).

\paragraph{Summary}

The complex prior model for multiple teachers meets all three of our modeling goals: stability of contextual variation in the face of coarticulation; stability of umlaut in the presence of categoricity bias; and rapid change in the population from stable contextual variation to stable umlaut as system parameters ($a$, $\lambda$) are varied around certain values.

\section{Discussion}
\label{discussionSec}

This paper has explored how assumptions about channel bias, categoricity bias, and population structure translated into population-level dynamics of a continuous parameter, evaluating models by their ability to meet two goals reflecting empirical cases of sound change: (1) the possibility of stable contextual variation and change to stable umlaut, in the presence of forces promoting the other outcome, and (2) a nonlinear transition from stable variation to sound change as a function of system parameters.

The first goal was met by all models where both a bias promoting change and a bias promoting stability were present: 
in both simple and complex prior settings, stable contextual variation can be maintained even in the presence of channel bias, and change to stable umlaut can occur even in the presence of categoricity bias. 
This is an important result, for two reasons related to the prevalence of both stable variation and sound change in the world's languages. First,
it shows that it is possible to develop a model of sound change involving channel bias that does not overapply 
\citep{weinreich1968empirical,baker2011variability,baker2008addressing}. Second, 
it shows that 
the distribution of a continuous parameter in the population does not necessarily come to reflect the structure of learners' hypothesis space, when other forces (such as channel bias) are present. Convergence to the prior has been emphasized in the cultural evolution literature \citep{griffiths2007language,kirby2007innateness,reali2009evolution,griffiths2013effects}, and would not allow for the possibility of \emph{both} stable variation and change to stable umlaut, when the prior reflects both possibilities. 
Instead, both stability and change are possible in a model where biases promoting each outcome are both present.


Our second goal concerned \emph{how} stability gives way to change as a function of the relative strength of these biases.  Models in which learners were equipped with a complex prior showed a bifurcation: change from one stable state (contextual variation) to another (umlaut) occurred suddenly as the relative strength of the biases favoring each stable state
is varied past a critical value, at which point `actuation' can be said to have occurred.  Bifurcations in linguistic populations have been suggested as a key mechanism underlying the actuation of linguistic change, but to our knowledge have previously only been shown to occur in models of change involving discrete variants \citep{komarova2001evolutionary,niyogi2006computational,niyogi2009proper,sonderegger2010combining}. Our demonstration that bifurcations are possible in a population of learners of a distribution of a continuous parameter supports the hypothesis that bifurcations play a key role in the actuation of language change more generally, and suggests that the ongoing empirical quantification of forces corresponding to channel and categoricity biases will be crucial to a detailed account of sound change actuation \citep{wilson2006learning,moreton2008analytic,sonderegger2010rational}.

Turning to the role of population structure,  we observed significant differences between single- and multiple-teacher settings. These differences are important
 given the prevalence of the single-teacher assumption in much of the sound change modeling literature \citep{kirby2013role,pierrehumbert2001exemplar,wedel2006exemplar}, and echo similar differences found in
previous work on the evolution of discrete linguistic traits \citep{dediu2009genetic,burkett2010iterated,niyogi2009proper,smith2009iterated}. For naive learners,
single-teacher scenarios result in ever-increasing population variance. In the simple prior cases, convergence to a form reflecting the prior was seen in single-teacher settings \citep{griffiths2013effects}, but not in multiple-teacher settings. For single teachers in the complex prior setting, the prior was reflected not in terms of individual's distribution of the learned phonetic parameter, but in terms of the population-level mixture: rather than a majority of individuals learning a phonetic parameter with a value intermediate between the 
prior endpoints, individuals tended to learn a value at one endpoint or the other, with the population consisting of a mixture of such individuals. 
This last result contrasts sharply with abundant sociolinguistic evidence showing that the distribution of linguistic traits in individuals tends to mirror that of their speech community \citep{labovprinciples,fruehwald2013phonological}.
Conversely, the results from multiple-teacher settings are consistent with the finding that social network ties can act as a conservative force promoting entrenchment \citep{milroy1980language}.  Overall, our results in single- versus multiple-teacher settings suggest that in addition to categoricity bias, population structure itself can play a role in promoting stability of existing phonetic categories.
While assuming one versus multiple teachers greatly affected the dynamics, it is important to point out our potentially unintuitive finding that models assuming \emph{any} number of teachers greater than one resulted in very similar dynamics.  Thus, exactly how 
population structure affects the distribution of a linguistic parameter over time requires further study.
Given the crucial role that social networks play in the propagation of language change \citep{labovprinciples,milroy1980language}, we are currently extending this framework to handle different population structures with more complex teacher-learner relations, including socially stratified variation.  Future work should also consider different types of  biases promoting stability and change, such as asymmetries in the extent of contact between members and in the social weighting of groups and variants.  These are some of many ways
in which our current framework can be extended to better match the complex reality of sound change.
However, even in the relatively simply model presented here, we have shown that a solution to the actuation problem is possible: understanding why a language changes, or fails to change, requires attention not only those forces promoting change, but their interplay with the forces constraining it.

\section*{Acknowledgements}
Portions of this work were presented at the 35th Annual Conference of
the Cognitive Science Society \citep{kirby2013model}, the Workshop on
Variation in the Acquisition of Sound Systems (2013), and the Workshop
on Sound Change Actuation (2013), and at colloquia at The Ohio State
University and the University of Pennsylvania.  We thank those
audiences as well as Dan Dediu, Chris Ahern, and Kenny Smith for their comments and suggestions.
MS was supported by grants from the Social Sciences and Humanities Research Council of Canada (\#430-2014-00018), the Canadian Foundation for Innovation (\#32451), and the Fonds de recherche du Qu\'{e}bec soci\'{e}t\'{e} et culture (\#183356).

\setlength{\bibsep}{5pt}

{\small
\renewrobustcmd{\disambiguate}[3]{#2~#3}
\bibliographystyle{apalike}
\bibliography{../cognitiveScience/cogsciArticle}
}

\newpage

\appendices


\section{Model}
\label{app:modelSec}

We first review the general setting presented in \emph{Model}
in the main paper. We assume the terminology and notation introduced
there, with some additions to be used in derivations (summarized in Table \ref{notation}).

\begin{table}
\centering
\fbox{
\begin{tabular}{lcl}
$t$ & {} & generation\\
$n$ & {} & number of training examples\\
$m$ & {} & number of teachers\\
$k_j$ & {} & number of examples drawn from $j$th teacher\\
$y_i$ & {} & $i$th example \\
$\bar{y}$ & {} & mean of all examples \\
$\bar{y}_j$ & {} & mean of examples drawn from $j$th teacher\\ 
$c$ & {} & mean of F1 of V$_{12}$ \\
$\lambda$ & {} & mean of channel bias\\
$\omega^2$ & {} & variance of channel bias\\
$\mu_a$ & {} & mean of F1 of V$_1$ \\
$\sigma_a^2$ & {} & variance of F1 of V$_{12}$ \\
$M$ & {} & size of population\\
$C^t$ & {} & Random variable for contextual variant in generation $t$\\
\end{tabular}
}
\caption{Notation.}
\label{notation}
\end{table}



Each generation at time $t$ consists of $M$ agents, who act as teachers for $M$ learners in generation $t+1$.  Each learner receives $n$ examples, drawn from $m$ teachers, with values $\vec{y} = (y_1, \ldots, y_n)$.  A new set of $m$ teachers from generation $t$ is drawn (with replacement) for each learner in generation $t+1$.  For a given learner, which teacher the $i$th example comes from is chosen randomly (each teacher has probability $1/m$), and $k_j$ denotes the number of examples received from the $j$th teacher, $\bar{y}_j$ the mean F1 of the examples received from the $j$th teacher, and $\bar{y}$ the mean F1 of all $n$ examples.

The distribution of V$_{12}$ for an agent with contextual variant $c$ is
\begin{equation} 
\text{V}_{12} \sim N(c, \sigma_a^2)
\label{app:threeDists}
\end{equation}
The random variable $C^t$ corresponds to the contextual variant used by members of generation $t$.  We use lower-case $c$ to refer to draws from this random variable, at times with subscripts ($c_j$ will refer to the contextual variant for the $j$th teacher) or a hat ($\hat{c}$ will refer to an individual learner's estimate of the contextual variant).  All productions of V$_{12}$ are subject to a channel bias with distribution $N(\lambda, \omega^2)$.  Thus, F1 for a teacher with contextual variant $c$ is distributed as
\begin{equation}
  F1 \sim N(c - \lambda, \sigma_a^2 + \omega^2)
  \label{app:v12ProdBias}
\end{equation}

We assume that $M$ is very large ($M \to \infty$), in which case the evolution of the distribution of $C^t$ is deterministic, and can be described by a dynamical system.  Analyzing the dynamical system under different assumptions lets us understand how different assumptions about bias and population structure affect the population-level distribution of the continuous phonetic parameter over time, analogously to existing dynamical systems models of language change which consider discrete linguistic variants (e.g.\ \citealp{mitchener2003bifurcation,niyogi2006computational,niyogi1997evolutionary,nowak2001evolution}).

It is worth briefly contrasting this setting with that considered in `iterated learning' (IL) models which are common in the language evolution literature, where each generation consists of a single member ($M=1$) (e.g.\ \citealp{kirby2007innateness,smith2003iterated,griffiths2007language,reali2009evolution,griffiths2013effects}).  In IL models, the state of the population is a stochastic process: it consists of a single value of $c$ at each time point, and can be described as a discrete-time Markov chain $c^t$. IL studies generally examine the evolution of this Markov chain: what would the distribution of values of $c^t$ be if the chain were iterated a large number of times?\footnote{\citet[][pp.\ 470--471]{griffiths2007language} do consider a \emph{continuous-time} population-level model as an extension of their discrete-time $M=1$ models, corresponding to a continuous linear dynamical system.  However, the vast majority of IL studies assume discrete generations of size 1.}  At time $t$ in any given iteration, there is only \emph{one} value of $c$.
In contrast, in our infinite-population setting we are examining the evolution of $C^t$, i.e.\ the \emph{distribution} of $c$ in the population at time $t$.  In other words, we are interested not in how a single parameter evolves (stochastically) over time, but in how the distribution of this parameter in a population evolves (deterministically) over time. A more detailed presentation of the difference between iterated learning and the `social learning' setting where $M = \infty$ is given by \citet{niyogi2009proper}.

\section{Naive learning models}
\label{app:naiveSec}

Here we derive all analytical results referred to in  \emph{Naive learning models}
 in the main paper.

We first consider maximum-likelihood (ML) learners who are ``naive'' in the sense of having no prior over estimates of $c$.  This setting is closely 
 related to  `blending inheritance' models of cultural evolution of a quantitative character presented by \citet[][71ff]{boyd1985culture}, which we make use of below.

\subsection{Naive learning models: single teacher}
\label{model11Sec}
\label{sec:singleTeacher}


Each learner in generation $t+1$ is associated with a value $c$ (one draw from 
 $C^t$, representing the single teacher's contextual variant), which is used to generate $n$ training examples for that learner.  Let $Y_1, \ldots, Y_n$ be the random variables corresponding to these examples, which take on values $\vec{y}=(y_1, \ldots, y_n)$, and let $\bar{y}$ be the mean of this sample.
%
Each example is normally distributed, following \eq{v12ProdBias}.
Because $Y_1, \ldots, Y_n$ are independent and normally distributed, their mean is also normally distributed, with the same mean and reduced variance:
\begin{equation}
f_{\frac{Y_1 + \cdots + Y_n}{n}}(\bar{y} \,|\, C^t = c) = {N}_{\bar{y}}(c - \lambda, (\sigma^2_a+ \omega^2)/n)
\label{meanY}
\end{equation}
Given $\bar{y}$, the learner's ML estimate of the contextual variant, assuming the data was generated by \eq{app:threeDists}, is $\hat{c}=\bar{y}$. Thus, using Eq.\ \ref{meanY}, the distribution over values of $\hat{c}$ the learner could acquire given $c$ is:
\begin{equation}
f_{C^{t+1}}(\hat{c} \,|\, C^{t} = c) = {N}_{\hat{c}}(c - \lambda, (\sigma^2_a + \omega^2)/n)
 \label{noisy}
\end{equation}
that is, $\hat{c}$ is a noisy version of $c$, decreased by the mean channel bias ($\lambda$).


We are interested in the evolution of the distribution of $c$: that is, the distribution of $C^{t+1}$ as a function of the distribution of $C^{t}$.  It is not in general possible to analytically derive what $f_{C^{t+1}}(c)$ is for an arbitrary $f_{C^{t}}(c)$.  However, we can get a sense of the evolution of the distribution of $c$
 by examining how its mean and variance change over time.

To do so, first consider the case where $\lambda = 0$.  The learner's estimate of $c$ can then be written as 
\begin{equation}
  \hat{c} = \sum_{i=1}^n \frac{1}{n} (c_i + \epsilon_i)
  \label{brEquiv1}
\end{equation}
where $c_i = c$ and $\epsilon_i \sim N(0, \sigma_a^2 + \omega^2)$.  In this form, our setting can be related directly to the classic `blending inheritance' model of a quantitative character \citep[][71ff]{boyd1985culture}, where:
\begin{itemize}
\item  A child in generation $t+1$ takes the mean value of the character from $n$ cultural parents (the $c_i$).
\item  Her observation of the $i$th cultural parent is distorted by a noise term ($\epsilon_i$).  
\item The distribution of $C^t$ is the distribution over cultural parents in generation $t$.
\end{itemize}
Having made this equivalence, Eqs.\ 3.21 and 3.22 of \citet{boyd1985culture} (rewritten using our notation) give the evolution of the mean and variance of $C^t$:
\begin{align}
  E[C^{t+1}] & = E[C^t]  \label{appEq1} \\
  \text{Var}[C^{t+1}] & = \sum_{i=1}^n \frac{1}{n^2}(\text{Var}[C^t] + \sigma_a^2 + \omega^2) + 2 \sum_{i=1}^{n-1}\sum_{j=i+1}^n  \frac{1}{n^2}(\text{Cov}(\epsilon_i, \epsilon_j) + 
  \text{Var}[C^t] \cdot \text{Corr}(c_i, c_j))
\label{appEq2}
\end{align}
In our case, $\text{Cov}(\epsilon_i, \epsilon_j)=0$ (because $\epsilon_i$ and $\epsilon_j$ are independent) and  $\text{Cor}(c_i, c_j)=1$ (because all the $c_i$ have the same value).
After some algebra, \eq{appEq2} thus simplifies to
\begin{equation}
  \text{Var}[C^{t+1}] = \text{Var}[C^t] + \frac{\sigma_a^2 + \omega^2}{n}
  \label{appEq3}
\end{equation}

\eq{appEq1} and \eq{appEq3} describe the evolution of the mean and variance of $C^t$ when $\lambda = 0$.  In the case where $\lambda > 0$, the learner's estimate of $\hat{c}$ changes by subtracting the constant $\lambda$ from the right-hand side of \eq{brEquiv1}, which entails subtracting $\lambda$ from the right-hand side of \eq{appEq1} and 0 from the right-hand side of \eq{appEq3}.\footnote{Because if $X$ is a random variable and $a$ is a constant, $E[X - a] = E[X] - a$ and $\text{Var}[X-a] = \text{Var}[X]$.}  The evolution of the mean and variance of $C^t$ are thus
\begin{align}
  E[C^{t+1}] & = E[C^t] - \lambda \label{app:mod11Mean} \\
  \text{Var}[C^{t+1}] & = \text{Var}[C^t] + \frac{\sigma_a^2 + \omega^2}{n} \label{app:mod11Var}
\end{align}
which are Eqs.\ \ref{mod11Mean}--\ref{mod11Var} in the main paper.


We note that although results from \citet{boyd1985culture} were used to
derive these evolution equations, the result that the variance of $c$
increases without bound over time (Eq.\ \ref{app:mod11Var}, illustrated in
Fig. \ref{vars}, panel 1) contrasts with their well-known finding that
blending inheritance in general \emph{reduces} variance of a
quantitative trait over time, as emphasized in their discussion (p.\ 75). However, stable or increasing variance
are possible for particular cases of \citeauthor{boyd1985culture}'s model, such as the case considered here where each learner has a single cultural parent and there is noise in estimating the parent's cultural model. 

\subsection{Naive learning models: multiple teachers}
\label{app:naiveMultipleSec}

We now consider the case where each learner in generation $t+1$ receives $n$ examples from $m>1$ teachers in generation $t$.  That is, values $c_1, \ldots, c_m$, corresponding to the $m$ teachers, are drawn i.i.d.\ from $C^t$, and the teacher who generates each example is chosen randomly (with replacement).  We assume that $n > 1$.\footnote{If $n=1$, the multiple-teacher case is the same as the single-teacher case already considered.}
Let $k_j$ denote the number of examples drawn from the $j$th teacher ($k_1 + \cdots + k_m = n$) , and let $\vec{k} = (k_1, \ldots, k_m)$.  Thus, 
$\vec{k}$ follows a multinomial distribution with $n$ trials and event probabilities $p_1, p_2, \ldots, p_m = 1/m$.  Without loss of generality, we can assume that examples $1, \ldots, k_1$ come from teacher 1, examples $k_1 + 1, \ldots, k_1 + k_2$ come from teacher 2, and so on.  Let $\bar{y}_j$ denote the mean of the examples from the $j$th teacher.
%
The learner's ML estimate, $\hat{c} = \bar{y}$, can then be rewritten as:
\begin{align}
  \hat{c} = \bar{y} &= \frac{1}{n} (y_1 + \cdots + y_n) \nn \\
  &  = \frac{1}{n} \sum_{j=1}^m k_j \bar{y}_j
  \label{appEq4}
\end{align}
Note that conditional on $k_j$, each $\bar{y}_j$ can be thought of as the learner's ML estimate of $c_j$ using $k_j$ examples from a \emph{single} teacher.  Thus, by the same logic used to derive Eqs.\ \ref{appEq1}, \ref{appEq3}, we have
\begin{align}
  E[\bar{y}_j] &= E[C^t] - \lambda \label{yiMean} \\
  \text{Var}[\bar{y}_j] &  = \text{Var}[C^t] + \frac{\sigma_a^2 + \omega^2}{n} \label{yiVar}
\end{align}
Because $\hat{c}$ is drawn from $C^{t+1}$, taking the expectation of \eq{appEq4} and substituting in \eq{yiMean} gives:
\begin{align}
  E[C^{t+1}] = E[\hat{c}] & = E[\frac{1}{n} \sum_{j=1}^m k_j \bar{y}_j] \nonumber \\
& = \sum_{\vec{k}} P(\vec{k}) E[\frac{1}{n} \sum_{j=1}^m k_j \bar{y}_j | \vec{k}] \nonumber \\
& = \sum_{\vec{k}} P(\vec{k}) \frac{1}{n} \underbrace{\sum_{j=1}^m k_j}_{=n} \underbrace{E[\bar{y}_j]}_{\text{Eq.\ \ref{yiMean}}} \nonumber \\
& = \sum_{\vec{k}} P(\vec{k}) (E[C^t] - \lambda ) \nonumber \\
& = E[C^t] - \lambda 
\end{align}

Similarly, taking the variance of \eq{appEq4} gives:
\begin{align}
\text{Var}[C^{t+1}] = \text{Var}[\hat{c}] & = \text{Var}[\sum_{j=1}^m \frac{k_j}{n} \bar{y}_j] \label{appEq41} \\
& = E_{\vec{k}}[\text{Var}[\sum_{j=1}^m \frac{k_j}{n} \bar{y}_j \, | \, \vec{k}]] 
- \text{Var}_{\vec{k}}[E[\sum_{j=1}^m \frac{k_j}{n} \bar{y}_j \, | \, \vec{k}]] \quad \text{(law of total variance)} \nonumber \\
& = E_{\vec{k}}[\sum_{j=1}^m \frac{k_j^2}{n^2} \underbrace{\text{Var}[\bar{y}_j]}_{\text{Eq.\ \ref{yiVar}}}] + \underbrace{\text{Var}_{\vec{k}}[E(C^t) - \lambda]}_{0} \nonumber \\
& = E[ \sum_{j=1}^m \frac{k_j^2}{n^2} (\text{Var}(C^t) + \frac{\sigma_a^2  + \omega^2}{k_j})] \nonumber \\
& = \frac{\text{Var}(C^t)}{n^2}\sum_{j=1}^{m} E[k_j^2] + \frac{\sigma_a^2 + \omega^2}{n} \underbrace{E[\sum_{j=1}^m \frac{k_j}{n}]}_{=1}
\label{appEq5}
\end{align}

Using the expressions for $E[k_j]$ and $\text{Var}[k_j]$ for a multinomial distribution (where $p_j$ is the probability of the $j^{\text{th}}$ outcome):
%
\begin{align}
  E[k_j^2] &= E[(k_j - \underbrace{E[k_j]}_{ = n p_j = \frac{n}{m}})^2] + E[k_j]^2  \nonumber \\
  & = \underbrace{\text{Var}[k_j]}_{ = n p_j (1-p_j) = n (\frac{1}{m})(1 - \frac{1}{m})}  + \frac{n^2}{m^2} \nonumber \\
  & = \frac{nm - n + n^2}{m^2}
\end{align}
Substituting into \eq{appEq5} gives
\begin{align}
\text{Var}[C^{t+1}] & = \frac{\text{Var}(C^t)}{n^2} m \cdot \frac{nm - n + n^2}{m^2} + \frac{\sigma_a^2 + \omega^2}{n} \nonumber \\
& =  \frac{\sigma_a^2 + \omega^2}{n} + \text{Var}[C^t] \frac{n + m - 1}{nm}
\label{appEq6}
\end{align}
The evolution equations of the mean and variance are then
\begin{align}
E[C^{t+1}] & = E[C^t] - \lambda  \label{mTeachersMean}\\
\text{Var}[C^{t+1}] & =  \frac{\sigma_a^2 + \omega^2}{n} + \text{Var}[C^t] \frac{n + m - 1}{nm} \label{mTeachersVar}
\end{align}
Thus, the mean of $c$ always decreases without bound, as in the single-teacher case (\eq{app:mod11Mean}), regardless of the number of teachers or the number of examples.

Turning to the variance, define $B = \frac{nm}{n+m-1}$. Because $m>1$ and $n>1$ (by assumption):
  \begin{align}
    (n-1)(m-1) > 0 & \implies n+m - 1 < nm \nonumber \\
    & \implies B > 1 \label{lessThan1Cond}
  \end{align}
  The variance evolution equation is an iterated map of the form 
  \begin{equation}
    x_{t+1} = K_1 +  x_{t}/B
\label{itMap}
  \end{equation}
where $K_1$ and $B$ are constants.
Because $|B| > 1$, the map has a unique fixed point $\alpha_*$ which it converges to from any starting point \citep{hirsch2004de}.  In particular, letting $\text{Var}[C^1]$ be the variance of $c$ in generation 1, we can rewrite \eq{mTeachersMean} as 
  \begin{equation}
    \text{Var}[C^{t}] = \alpha_* + \frac{(\text{Var}[C^1] - \alpha_*)}{B^{t-1}}
    \label{varDist}
  \end{equation}
where
\begin{equation}
  \alpha_* = \frac{m}{m-1} \frac{\sigma_a^2 + \omega_a^2}{n-1}
\end{equation}
is the fixed point.

Thus, for multiple teachers, the variance quickly converges to a fixed point $\alpha_*$, with its distance from $\alpha_*$ decreasing geometrically (\eq{varDist}, illustrated in Fig. \ref{vars}, panels 2--3).  The value of the stable variance decreases as the number of examples ($n$) or the number of teachers ($m$) is increased.  For example, for the two-teacher case ($m=2$), \eq{varDist} gives
\begin{equation}
  \text{Var}[C^t] \to \frac{2 (\sigma_a^2 + \omega^2)}{n-1} \quad \text{(two teachers)}
\end{equation}
which is Eq.\ \ref{mod13Var} in the main paper.  For the case where learners learn from all teachers ($m=M$), in the limit considered in our setting where $M \to \infty$, \eq{varDist} gives
\begin{equation}
\text{Var}[C^t] \to \frac{\sigma_a^2 + \omega^2}{n-1} \quad \text{(all teachers)}
\end{equation}
which is Eq.\ \ref{mod12Var} in the main paper.



\section{Simple prior models}
\label{app:simplePriorSec}

Here we derive all analytical results referred to in \emph{Simple prior
models}
in the main paper.

In these models, we again assume (as in the naive learning models) that a learner in generation $t+1$ estimates the mean of the contextual variant based on the assumption that her data (from generation $t$) is generated i.i.d.\ from a gaussian source with a fixed $c$ (\eq{app:threeDists}). However, we now assume that she 
  has a gaussian prior over how likely different values of $c$ are: 
  \begin{equation}
    f_{C^{t+1}}(c) = {N}_c(\mu_a, \tau^2)
\label{simplePriorEq2}
 \end{equation}
which is updated to a posterior distribution based on the data ($f_{C^{t+1}}(c \,|\, \vec{Y} = \vec{y})$).\footnote{This learning algorithm is similar to that considered by \citet{griffiths2013effects} in a study of the evolution of a continuous parameter, but their iterated learning setting (where $M=1$) differs from the population setting considered here (where $M$ is large),  as discussed above. 
We compare our results to theirs below.}



In this setting, the learner can be seen as performing a particularly simple case of Bayesian linear regression (see e.g.\ \citealp{bishop2006pattern}). 
where she is finding the constant ($c$) that best matches the mean of the data ($\vec{y}$) in the least-squares sense, and there is a gaussian prior on possible values of $c$.  The gaussian prior is the conjugate prior, so the posterior distribution of $c$ is gaussian as well. 
Using Eq.\ 3.49--3.51 from \citet{bishop2006pattern}, the posterior can be shown to be:\footnote{In particular, by making these substitutions into Eq.\ 3.49--3.51: $\mathbf{w} = (\vec{c})$, $\mathbf{S_0} = (\tau^2)$, $\beta = 1/\sigma_a^2$, $\mathbf{\Phi} = (\overbrace{1, \ldots, 1}^{n\text{ times}})^T$, $\mathbf{m_0} = \mu_a$.}
\begin{equation}
f_{C^{t+1}}(c \, | \, \vec{Y} = \vec{y}) = 
N_{c}(\frac{\bar{y} + \mu_a \frac{\sigma_a^2}{n\tau^2}}{1 + \frac{\sigma_a^2}{n\tau^2}},
\frac{\sigma_a^2}{n} \frac{1}{1 + \frac{\sigma_a^2}{n\tau^2}} )
\label{simplePriorEq3}
\end{equation}

The learner must pick a point estimate of the contextual variant to use for generating training data for the next generation. The two common ways of obtaining a point estimate from a posterior distribution are taking the maximum a-posteriori value or the expected value.  These are equivalent for \eq{simplePriorEq3} (because the mean and mode of a normal distribution are identical), namely:
\begin{equation}
\hat{c} = \frac{\bar{y} + (D-1) \mu_a}D
\label{simplePriorEq4}
\end{equation}
where we 
abbreviate the denominator of \eq{simplePriorEq4} as
\begin{equation}
  D = 1 + \frac{\sigma_a^2}{n\tau^2}.
  \label{simplePriorEq5}
\end{equation}

Using the same notation as above (Table \ref{notation}),
we now determine the evolution of the mean and variance of $C^t$ for a population of simple prior learners whose estimate of $c$ is given by \eq{simplePriorEq4}.   To reduce the number of cases which need to be considered below, we assume that $n>1$, $\sigma_a > 0$, and $\tau > 0$: that is, each learner receives more than one example, there is some variability among a speaker's productions of V$_{12}$, and the categoricity prior is not infinitely strong.

\subsection{Simple prior models: single teacher} 
\label{singleTeacherSec}
For the case where $m = 1$, the distribution of $\bar{y}$ is still given by \eq{meanY}, where $c$ is the value of the contextual variant used by the single teacher. Because $\mu_a$ and $D$ are constants, using \eq{simplePriorEq4}, the distribution of $\hat{c}$ is then
\begin{equation}
f_{C^{t+1}}(\hat{c} \,|\, C^t=c) = N_{\hat{c}}(\frac{c - \lambda + (D-1)\mu_a}{D}, \frac{\sigma_a^2 + \omega^2}{nD^2})
\end{equation}
Examining \eq{noisy}, we see that if $X$ denotes the estimate of $\hat{c}$ (given the teacher's value of $c$) in the single-teacher naive learner case, then $\hat{c}$ for the current case is simply $X$ translated and divided by constants: $(X + (D-1)\mu_a)/D$.
Thus, \eq{app:mod11Mean} can be used to find the evolution of the mean:
\begin{align}
E[C^{t+1}] = E[\hat{c}] & = E[\frac{X + (D-1)\mu_a}{D}] \nn \\
& = \frac{1}{D}(E[C^t] - \lambda + (D-1)\mu_a)
\label{simplePriorSingleMean}
\end{align}
and \eq{app:mod11Var} can be used to find the evolution of the variance:
\begin{align}
\text{Var}[C^{t+1}] = \text{Var}[\hat{c}] & = \text{Var}[\frac{X+(D-1)\mu}{D}] \nn \\
& = \frac{\text{Var}[X]}{D^2} \nn \\
& = \frac{\sigma_a^2 + \omega^2}{nD^2} + \frac{\text{Var}[C^t]}{D^2}
\label{simplePriorSingleVar}
\end{align}

Now, note that the assumption that $\sigma_a, \tau > 0$ means that $D>1$, so that both \eq{simplePriorSingleMean} and \eq{simplePriorSingleVar} are iterated maps of the form in \eq{itMap} with $|B| > 1$. These maps have unique stable fixed points, thus, 
both the mean and the variance of $C^t$ rapidly converge to fixed points from any starting values.  Solving for the fixed points gives:\footnote{E.g.\ by setting $E[C^{t+1}]$ and $E[C^t]$ to $x$ in \eq{simplePriorSingleMean} and solving for $x$.}
\begin{align}
  E[C^t] & \to \mu_a - \lambda n \frac{\tau^2}{\sigma_a^2} \label{simplePriorSingleMeanLim} \\
  \text{Var}[C^{t+1}] & \to \tau^2 \frac{(1 + \frac{\omega^2}{\sigma_a^2})}{(2 + \frac{\sigma_a^2}{n \tau^2})}
  \label{simplePriorSingleVarLim}
\end{align}
\eq{simplePriorSingleMeanLim} is Eq.\ \ref{simpleMean} in the main paper.  The mean of the contextual variant in the population converges to the value favored by the prior ($\mu_a$), minus an offset which depends on $\lambda$, $n$, $\tau$, $\sigma_a$ in intuitive directions: stronger net channel bias ($\lambda$) over the $n$ examples results in lower $c$, while stronger categoricity bias relative to the amount of production variability ($\tau/\sigma_a$) results in $c$ nearer to $\mu_a$.  

We discuss the expression for the stable variance below, along with the equivalent expression for the multiple-teacher case.

\subsubsection{Comparison with previous work}

Because our single-teacher simple prior scenario is particularly close to one of the iterated learning scenarios considered by \citet{griffiths2013effects},  it is worth comparing our results to theirs to see to what extent they diverge.\footnote{\citeauthor{griffiths2013effects} in fact mention `the value of a specific formant of a phoneme' as a motivating case (p.\ 955).}
%
Individual learners in their `category defined on a single dimension'
setting (pp.\ 954--956) learn in essentially the same way as our
single-teacher simple prior learners, except that no production bias
is applied. In addition, each generation consists of one
teacher/learner ($M=1$), compared to our $M = \infty$. Thus, the value
of $c$ at each time point is a Markov chain, which we write as $c^t$.
In our notation, \citeauthor{griffiths2013effects} show that (p.\ 966,
Eq.\ 11) 
\begin{equation} 
c^t \, | \, c^1 \sim N(\mu_a + c^1/D^{t-1}, \tau^2 \frac{(1 +
  \frac{\sigma_a^2}{n \tau^2})}{1 - D^{-2(t-1)}}) 
\label{griffithsEq1} 
\end{equation}
Although each generation in an iterated learning model consists of only one agent, there is a natural interpretation of $c^t$ in a population context (where $M = \infty$)
as describing the distribution of $C^t$ in a population of teacher/learners, each of whom learn from exactly one agent (in generation $t-1$) and teach exactly one agent (in generation $t$) (as pointed out by \citealp{niyogi2009proper,griffiths2013effects}).  (In other words, the population consists of an infinite number of iterated-learning chains run in parallel.)  This setting is slightly different from our single-teacher case (Fig 1, left panel in the main text), where two members of generation $t$ could have the same teacher (and some members of generation $t-1$ might never serve as teachers).  How does this slight difference affect the dynamics?  We can compare the stable state of the distribution of $C^t$ in the two cases by setting $\lambda = 0, \omega = 0$ in Eq.\ \ref{simplePriorSingleMeanLim}--\ref{simplePriorSingleVarLim}, and taking the limit $t \to \infty$ in \eq{griffithsEq1}:
\begin{align}
  E[C^t] &\to \mu_a \quad \text{(both models)} \\
  \text{Var}[C^t] & \to 
  \begin{cases}
    \tau^2 (2 + \frac{\sigma_a^2}{n \tau^2})^{-1} & \text{(our setting)} \\
    \tau^2 (1 + \frac{\sigma_a^2}{n \tau^2})   & \text{(iterated learning)}
  \end{cases}
\end{align}
Thus, the distribution of the coarticulation parameter comes to reflect the prior in both cases: the mean converges to the mean of the prior (in both models), while the variance converges to a value related to $\tau^2$ (the width of the prior), but which is smaller in our model than in the iterated learning model, by at least a factor of 2 (depending on the values of $\tau$, $\sigma_a$, and $n$).   The long-term dynamics are therefore similar in the two models, but slightly different.


\subsection{Simple prior models: multiple teachers}

When $m > 1$, we can proceed similarly to the naive-learner multiple teacher case,
defining $k_j$, $\bar{y}_j$, etc.\ in the same way.
The learner's point estimate of $\hat{c}$ is still given by \eq{simplePriorEq4}, which can be used to rewrite $\hat{c}$ as:
\begin{align}
\hat{c} = \frac{(D-1)\mu_a}{D} + \frac{\bar{y}}{D} & = \frac{(D-1)\mu_a}{D} + \frac{1}{nD} (y_1 + \cdots + y_n) \nn \\
& = \frac{(D-1)\mu_a}{D} + \frac{1}{nD} \sum_{j=1}^m k_j \bar{y}_j 
\label{simplePriorMultipleEq1}
\end{align}
As in \emph{Naive learning models: single teacher},
$\bar{y}_j$ can be thought of as the ML estimate made by a naive learner (in generation $t+1$) based on drawing $k_j$ examples from a single teacher in generation $t$.  Also, note that $D$ does not depend on any $k_j$.  Because $\hat{c}$ is drawn from $C^{t+1}$, taking the expectation of both sides of \eq{simplePriorMultipleEq1} gives:
\begin{align}
  E[C^{t+1}] = E[\hat{c}] & = \frac{(D-1)\mu_a}{D} + E[\frac{1}{nD} \sum_{j=1}^m k_j \bar{y}_j] \nonumber \\
& = \frac{(D-1)\mu_a}{D} + \sum_{\vec{k}} P(\vec{k}) E[\frac{1}{nD} \sum_{j=1}^m k_j \bar{y}_j | \vec{k}] \nonumber \\
& = \frac{(D-1)\mu_a}{D} + \sum_{\vec{k}} P(\vec{k}) \frac{1}{nD} \underbrace{\sum_{j=1}^m k_j}_{=n} \underbrace{E[\bar{y}_j]}_{\text{Eq.\ \ref{yiMean}}} \nonumber \\
& = \frac{(D-1)\mu_a}{D} + \sum_{\vec{k}} \frac{P(\vec{k})}{D} (E[C^t] - \lambda ) \nonumber \\
& = \frac{E[C^t] - \lambda  + (D-1)\mu_a}{D} \label{simplePriorMultipleMean}
\end{align}
Thus, the evolution of the mean in the multiple-teacher case
(\eq{simplePriorMultipleMean}) is the same as in the single-teacher
case (\eq{simplePriorSingleMean}).  In particular, the mean converges
to the value given in \eq{simplePriorSingleMeanLim}, which gives Eq.\ \ref{simpleMean} in the main paper.

Similarly, taking the variance of \eq{simplePriorMultipleEq1} gives
\begin{align}
\text{Var}[C^{t+1}] = \text{Var}[\hat{c}] & = \text{Var}[\frac{(D-1)\mu_a}{D} + \sum_{j=1}^m \frac{k_j}{nD} \bar{y}_j] \nonumber \\
& = \text{Var}[\sum_{j=1}^m \frac{k_j}{nD} \bar{y}_j] \\ 
& = \frac{1}{D^2} \underline{\text{Var}[\sum_{j=1}^m \frac{k_j}{n} \bar{y}_j]}
\end{align}
The underlined term is the same as \eq{appEq41} in the naive learner multiple-teacher case, and its derivation proceeds identically from that point on (up to \eq{appEq6}), to give
\begin{equation}
\text{Var}[C^{t+1}] = \frac{\sigma_a^2 + \omega^2}{nD^2} + \text{Var}[C^t] \frac{n + m - 1}{nmD^2}
\label{simplePriorMultipleVar1}
\end{equation}
Recall that for multiple teachers ($m>1$), provided that $n>1$ (which is true, by assumption), we have that $|(n+m-1)/nm| < 1$ (\eq{lessThan1Cond}).  Thus, because $D \ge 1$ as well (\eq{simplePriorEq5}), the evolution equation for the variance (\eq{simplePriorMultipleVar1}) is an iterated map of the form in \eq{itMap}, with $|B| > 1$, which has a unique stable fixed point.  Solving for it gives:
\begin{equation}
\text{Var}[C^{t+1}] \to \frac{\tau^2(1 + \frac{\omega^2}{\sigma_a^2})}{\frac{(n-1)(m-1)}{m}\frac{\tau^2}{\sigma_a^2} + (2 + \frac{\sigma_a^2}{n\tau^2})}
\label{simplePriorMultipleVar2}
\end{equation}
Comparing \eq{simplePriorMultipleVar2} with \eq{simplePriorSingleVar},
we see that the stable variance decreases monotonically as the number
of teachers ($m$) is decreased, when all other parameters are held
constant. (This fact is referred to in \emph{Naive learning models} in the main paper.)
In particular, the stable variances for the three values of $m$ considered in the main paper (1, 2, $\infty$) are:
\begin{align}
\text{Var}[C^{t+1}] & \to \tau^2 \frac{(1 + \frac{\omega^2}{\sigma_a^2})}{ (2 + \frac{\sigma_a^2}{n\tau^2})}
\quad &\text{(1 teacher)} \label{stabVar1} \\
& \to \frac{\tau^2(1 + \frac{\omega^2}{\sigma_a^2})}{\frac{(n-1)}{2}\frac{\tau^2}{\sigma_a^2} + (2 + \frac{\sigma_a^2}{n\tau^2})} & \text{(2 teachers)} \label{stabVar2} \\
& \to \frac{\tau^2(1 + \frac{\omega^2}{\sigma_a^2})}{(n-1) \frac{\tau^2}{\sigma_a^2} + (2 + \frac{\sigma_a^2}{n\tau^2})}
\quad & \text{(all teachers)} \label{stabVarInf}
\end{align}

These expressions for the stable variance are hard to understand intuitively.  We can get a sense of their behavior by taking $n$ to be large, in accordance with the intuition that each learner will receive many examples of a given phonetic category.   Taking the Taylor expansions of Eqs.\ \ref{stabVar1}--\ref{stabVarInf} in terms of $1/n$ gives:
\begin{align}
\text{Var}[C^{t+1}] & \to \tau^2 \frac{(1+\frac{\omega^2}{\sigma_a^2})}{2} - \sigma_a^2 \frac{(1+ \frac{\omega^2}{\sigma_a^2})}{4n} + O(\frac{1}{n^2}) \quad & \text{(1 teacher)}\\
& \to \sigma_a^2 \frac{2(1 + \frac{\sigma_a^2}{\omega^2})}{n} + O(\frac{1}{n^2}) \quad & \text{(2 teachers)} 
\label{simplePriorVarLimExpansion2}
\\
& \to \sigma_a^2 \frac{(1 + \frac{\sigma_a^2}{\omega^2})}{n} + O(\frac{1}{n^2}) \quad & \text{(all teachers)}
\end{align}
where $O(\frac{1}{n^2})$ denotes a constant divided by $n^2$.
(These are Eqs.\ \ref{simpleVar1}--\ref{simpleVar3} in the main paper.)  Thus, there are two important differences between the form of the stable variance for $m=1$ and $m>1$:\footnote{In general, for $m > 1$ teachers, the 2 in \eq{simplePriorVarLimExpansion2} is replaced by $m/(m-1)$.}
\begin{itemize}
\item  First, the stable variance for $m=1$ always reflects the prior (in the sense that the expression involves $\tau$), for any $n$, while the stable variance for $m>1$ does not ($\tau$ only enters into second-order terms).
\item  Second, the stable variance for $m>1$ goes to 0 as $n$ is increased, while the stable variance for $m=1$ goes to a constant value which reflects the prior. Thus, a population of simple prior learners who receive many examples would eventually show \emph{no} variability in their contextual variants (values of $c$) for two or more teachers, while the same population with single-teacher learning would show variability in their contextual variants.
\end{itemize}
%

\section{Complex prior models}
\label{app:complexPriorSec}

Here we provide a more detailed description of the complex prior
models, and technical details of the complex prior model simulations
whose results are given in \emph{Complex prior models} in the main paper.

In these models, we again assume (as in the Simple Prior models) that learners estimate the mean of the contextual variant based on the assumption that data is generated i.i.d.\ from a gaussian source with a fixed $c$, and that they have a prior over how likely different values of $c$ are, which is now given by:
\begin{equation}
  f_{C^{t+1}}(c) \propto \left[ a(\mu_a - \mu_i)^2 + (c - (\mu_a + \mu_i)/2)^2 \right]
  \label{app:post3}
\end{equation}
(We write $\propto$ instead of $=$ because $f_{C^{t+1}}(c)$ must be scaled by
some constant to be a probability distribution.)
 The strength of this prior is controlled by $a$: as $a \to 0$, values
 of $c$ near $\mu_a$ and $\mu_i$ are maximally preferred relative to
 values in between (Fig.\ \ref{simplePriorFig}A in the main paper).

This prior is updated to a posterior distribution based on the data $\vec{y}$. The log of the posterior is given by:
\begin{equation}
\log(f_{C^{t+1}}(c | \vec{y}))  = -\sum_{i=1}^{n} \frac{(y_i - c)^2}{2 \sigma_a^2} +
\log [  a(\mu_a - \mu_i)^2 + (c - (\mu_a + \mu_i)/2)^2 ]+ \text{constant}
\label{post4}
\end{equation}
where the constant is a term which does not depend on $c$.

We assume that each learner takes the MAP estimate of $\hat{c}$ over
the interval $[\mu_i, \mu_a]$ based on this posterior. Because this
MAP estimate is not in general possible to compute analytically, it is
also not possible to obtain analytical expressions for the evolution
of the mean and variance of $C^t$, as in the naive learner and simple
prior models.  Thus, we proceeded by simulation to examine the evolution of $C^t$.  

\subsection{Simulations: setup}
\label{simulationsSetupSec}


As an approximation to the deterministic evolution which would result
for $M = \infty$, we carried out simulations using $M = 50000$ for the
single-teacher setting and $M = 2500$ for the multiple-teacher
settings.  These values were large enough to give behavior very close
to deterministic for the multiple-teacher settings, and roughly
deterministic behavior in the single-teacher setting.  In the
single-teacher setting, it was not possible to obtain effectively
deterministic behavior for any feasible value of $M$. This should be
kept in mind when examining the results of the single-teacher
simulations, where there is a small stochastic component to the
results (relative to $M = \infty$), compared to the multiple-teacher
simulations, where the results approximate the $M = \infty$ case very
closely.

In each simulation run, all parameters except $a$, $\lambda$, and $m$
(the number of teachers) were set to the same values: $\mu_a = 730$,
$\mu_i = 530$, $\sigma_a = 50$, $n = 100$.  Runs were conducted for
values of $a \in [0.001, 0.05]$ and $\lambda \in [0, 2.0]$, for the
single-teacher, two-teacher, and $M$-teacher cases ($m=1, 2, M$).
Each  run
began by
assigning a value of $c$ to the $M$ learners in generation 1, drawn
according to a starting distribution. Because we are primarily
interested in the evolution of a population which begins with the
contextual variant uniformly pronounced similarly to V$_1$, we always
used $C^1 \sim N(\mu_a - 10, 10)$ as the starting distribution.  For
each of the $M$ learners in generation $t$, where $t>2$, $m$ teachers
were drawn at random from generation $t-1$, and used to generate $n =
100$ examples (with the teacher for each example chosen randomly from
the $m$ teachers, with replacement).  The learner's MAP estimate
$\hat{c}$ for this data was found by maximizing \eq{post4} over values
of $c \in [\mu_i, \mu_a]$, using the unidimensional
\texttt{optimize()} function in \texttt{R}, which uses ``a combination
of golden section search and successive parabolic interpolation''
\citep{R}.\footnote{\texttt{optimize} is guaranteed to find the global
  maximum only if \eq{post4} is unimodal over the interval $c \in
  [\mu_i, \mu_a]$; otherwise, it is only guaranteed to find a local
  maximum. Whether \eq{post4} is unimodal over this interval in
  general depends on the values of the data ($\vec{y}$) and the system
  parameters ($a$, $\sigma_a$, etc.).  \eq{post4} can be shown to be
  concave on $[\mu_i, \mu_a]$ for \emph{any} $\vec{y}$ if the
  condition $a > \frac{4 \sigma_a^2}{n(\mu_a - \mu_i)^2}$ holds, which
  is the case for almost all simulation runs considered here (those
  with $a > 0.0025$).  Note that concavity is not a necessary
  condition for \texttt{optimize} to find the global optimum, which it
  seems to nearly always do anyway in our setting.  We satisfied
  ourselves that \texttt{optimize} getting stuck in local maxima was
  not a problem by comparing the results with those obtained by using
  grid search instead, for a subset of the runs.}  

 For the two and $M$-teacher cases, simulations were run until $t=2500$, at which point the distribution of $C^t$ had always reached a stable state (by visual inspection). For the single-teacher case, which converged much more slowly, simulations were run until the mean, 5th percentile, and 95th percentile of the distribution of $C^t$ had (each) not changed by more than 2 in 500 generations. If this criterion was not met by $t=10000$, the simulation was stopped.  At this point the distribution of $C^t$ had reached a stable state for runs corresponding to the dark red and dark blue regions of Fig.\ \ref{complexPriorFig}B Panel 1, though not necessarily for runs corresponding to the region in between.

\end{document}